# A Polynomial Time Algorithm For Finding
# Bayesian Probabilities from Marginal Constraints

John W. Miller and Rodney M. Goodman
Dept. of Electrical Engineering (116-81)
*California Institute of Technology*
Pasadena, CA 91125

## Abstract

A method of calculating probability values from a system of marginal constraints is presented. Previous systems for finding the probability of a single attribute have either made an independence assumption concerning the evidence or have required, in the worst case, time exponential in the number of attributes of the system. In this paper a closed form solution to the probability of an attribute given the evidence is found. The closed form solution, however does not enforce the (non-linear) constraint that all terms in the underlying distribution be positive. The equation requires $\mathcal{O}(r^3)$ steps to evaluate, where $r$ is the number of independent marginal constraints describing the system at the time of evaluation. Furthermore, a marginal constraint may be exchanged with a new constraint, and a new solution calculated in $\mathcal{O}(r^2)$ steps. This method is appropriate for calculating probabilities in a real time expert system.

## Introduction

The problem of calculating a single probability value from a set of evidence is of great importance to the field of artificial intelligence. This paper addresses this problem in the case where the probability is to be calculated from a set of marginal constraints. Previous methods have either made a conditional independence (CI) assumption concerning the evidence, or have required time exponential in the number of attributes which must be considered simultaneously. When the evidence is in the form of a set of rules firing, the form of the rules themselves often force the evidence to be conditionally dependent. Several important methods which do not make the CI assumption have been proposed. Konolige described a system which dealt with the problem by subdividing the evidence into Local Event Groups (LEGS) which are then updated [Konolige, 1979]. Within the LEGs the updating is exponential in the number of attributes, but the exponent is kept

John Miller is supported by NSF grant no. ENG–8711673

small. Since the LEGS may overlap (share attributes) the problem still remains in finding a consistent update for all of the LEGS. In some cases the overlap of the LEGS may form a tree structure, so that the LEGS can be updated in an order which insures consistent solutions.

A second important method for the correct updating of probabilities was described by Cheeseman [Cheeseman, 1983]. This method uses Lagrange multipliers to find a set of non-linear equations which can be solved to find the desired probability. The number of simultaneous equations is non-exponential in the number of attributes. Each equations, however, has length which is exponential in the number of attributes in the worst case. Cheeseman shows that typically these equations can be greatly simplified so that the combinatoric explosion is controlled.

This paper presents another method which is shown to be non-exponential in the number of attributes. An important feature of this algorithm is the output probability is found in a closed form equation which does not require iterative methods of solution. This is a requirement for real-time expert systems which must know precisely how long the probability calculation will take.

## Inference with a Set of Rules

A probabilistic rule is a statement of the form

$$\mathbf{Y} = y \Rightarrow \mathbf{X} = x \ \text{ with probability p}$$

where upper case letters represent attributes and lower case letters represent attribute values. A rule here is taken to be a description of the joint probability distribution of the attribute value y with the attribute $\mathbf{X}$. It is called a *rule* because the information is only to be used in evaluating evidence in which it is known $\mathbf{Y} = y$. Fitting the concept of a logical rule, the evidence $\mathbf{Y} \neq y$ together with the rule $\mathbf{Y} = y \Rightarrow \mathbf{X} = x$ is not taken to imply anything about $\mathbf{X}$.

The following presentation assumes the attributes $F_1, F_2, \ldots F_n$ are binary inputs and a single attribute $\mathbf{X}$ is the desired output probability to be estimated. The development can be extended to general non-binary discrete attributes (although with much more complex notation). The evidence presented to the inference system $e = \{F_1 = f_1, F_2 = f_2, \ldots F_n = f_n\}$ is abbrieviated $e = \{f_1, f_2, \ldots f_n\}$. Similarly $F_1 \neq f_1$ is abbrieviated $\overline{f}_1$. The inference problem is to find $p(x|e)$. Using Bayes' rule, $p(x)$, and $p(\overline{x})$, the value of $p(x|e)$ may be calculated from $p(e|x)$ and $p(e|\overline{x})$.

$$\frac{p(x|e)}{1 - p(x|e)} = \frac{p(e|x)p(x)}{p(e|\overline{x})p(\overline{x})} \tag{1}$$

The following will demonstrate how to find $p(e|x)$. The same procedure may be used to find $p(e|\overline{x})$. Equation (1) may then be used to find $p(x|e)$. Note, although the following analysis involves equations of length exponential in n, the final result will not be exponential in $n$.

Let $z$ be the vector with length $2^n$ of conditional probabilities describing the records in which $\mathbf{X} = x$.

$$z = \begin{pmatrix} p(\overline{f}_1, \overline{f}_2, \ldots \overline{f}_n | x) \\ p(\overline{f}_1, \overline{f}_2, \ldots f_n | x) \\ \vdots \\ p(f_1, f_2, \ldots \overline{f}_n | x) \\ p(f_1, f_2, \ldots f_n | x) \end{pmatrix}$$

As stated previously, a rule describes the joint distribution of the left hand side (LHS) and the right hand side $\mathbf{X}$. This joint distribution can be viewed as a constraint on $z$. For rules $i = 1, 2, \ldots r$ we have the constraint

$$\alpha_i^T z = \beta_i$$

where $\alpha_i$ is a vector of length $2^n$. For example, let the number of attributes $n = 3$, and let rule number 1 be $F_1 = f_1 \Rightarrow \mathbf{X} = x$. This rule (when $F_1 = f_1$) provides the equation

$$\sum_{f_1, f_2} p(f_1, f_2, f_3 | x) = p(f_1 | x)$$

which is a constraint on $z$:

$$\alpha_1^T z = \beta_1$$
$$\text{where} \quad \alpha_1 = (\,0\,0\,0\,0\,1\,1\,1\,1\,)^T \quad \text{and} \quad \beta_1 = p(f_1 | x).$$

A rule is said to *fire* when the LHS attribute values are found in the evidence. The set of firing rules together create the constraint equation on $z$

$$Az = b \tag{2}$$

where $A$ is an $r$ by $2^n$ composite matrix ($r$ is the number of rules firing),

$$A = (\,\alpha_1 \alpha_2 \ldots \alpha_r\,)^T$$

and

$$b = (\,\beta_1 \beta_2 \ldots \beta_r\,)^T$$

Since $z$ represents a set of probabilities which sum to one, the $r$th constraint is always set to be $\alpha_r = (1\,1\,1\,\ldots 1)^T$, $\beta_r = 1$. The solution must also satisfy

$$z_i \geq 0 \quad (\text{for } i = 1, 2, \ldots 2^n) \tag{3}$$

In general, the $z$ vector is underconstrained by equations (2) and (3). The principal of least information says the best solution satisfying the constraints is the one which minimizes

$$I(z) = 2^n + \sum_{i=1}^{2^n} z_i log(z_i).$$ (4)

The function $I(z)$ has the following properties:

1. $I(z)$ is convex.
2. It is symmetric in its arguments.
3. It achieves its minimum at the equiprobable distribution.
4. It achieves its maxima at the $(0\ 0 \ldots 0\ 1\ 0\ 0 \ldots 0)$ distributions.

The principal of least information is well founded as a method of choosing a probability distribution [Lewis, 1959][Konolige, 1979]. The idea is that the confidence in the model decreases with the distance from the *apriori* probability distribution of $z$. Using the equiprobable solution $z_i = 2^{-n}$ (for $i = 1, 2, \ldots 2^n$) as the *apriori* distribution and defining the distance between distributions to be the minimum discriminant information measure from information theory leads to (4) [Lewis, 1959][Konolige, 1979].

To restate the problem:

$$Az = b, \quad z \geq 0, \quad I(z) \text{ minimized}$$ (5)

This may be solved using iterative techniques such as the convex simplex method [Ku & Kullback, 1969], or by using Lagrange multipliers [Cheeseman, 1983]. Both methods require worse case evaluation time exponential in $n$.

Let $l = 2^n$ and $y_i = z_i - l^{-1}$ for $i = 1, \ldots l$

Minimum $I(z)$ is equivalent to maximum entropy $H(z)$:

$$H(z) = -\sum_{i=1}^{l} z_i log(z_i) = -\sum_{i=1}^{l} (l^{-1} + y_i) log(l^{-1} + y_i)$$

Take the first two terms of the Taylor expansion (about $y_i = 0$) of the entropy function:

$$H(z) \approx log(l) - \sum_{i=1}^{l} (l^{-1} + y_i)(ly_i - .5l^2 y_i^2)$$

$$H(z) \approx log(l) - .5l||y||^2 + \sum_{i=1}^{l} (l^{-1} + y_i)$$ (6)

Now, consider the Euclidean norm square of $z$

$$||z||^2 = \sum_{i=1}^{l} z_i^2$$

$$||z||^2 = l^{-1}(1 + l\sum_{i=1}^{l} y_i^2)$$

Again taking the first two terms of the Taylor expansion:

$$-log||z||^2 \approx log(l) + l||y||^2 \qquad (7)$$

From equations 6 and 7 we get the approximation:

$$\text{Maximum } H(z) \approx \text{Maximum } .5(log(l) - log(||z||^2)) \equiv \text{Minimum } ||z||$$

Thus the approximation leads to the familiar estimation condition of Least Mean Square Distance. Replace $I(z)$ in equation (5) with $||z||$, the Euclidean norm. For probability distributions $||z||$ has all of the properties $1 - 4$. The problem described in (5) is now approximated by:

$$Az = b, \quad z \geq 0, \quad ||z|| \text{ minimized} \qquad (8)$$

It is difficult to estimate how the inference system performance will be affected by this approximation. In comparing two distributions $z$ and $y$ the approximation claims

$$sign(I(z) - I(y)) = sign(||z|| - ||y||) \quad \text{with high probability}$$

This hypothesis was tested empirically with the following series of tests. Two random distributions, $z$ and $y$, of length $l$ were created. (To make a random distribution, create uniformly distributed $[0,1)$ random elements, calculate their sum, and divide each element by this sum.)

Set test = agreement if $sign(I(z)-I(y)) = sign(||z||-||y||)$. The probability of agreement was calculated by computer simulation for value of $l$ from $2^2 - 2^{14}$. The probability of agreement was found to be $\approx 93\%$. This is not intended as a proof. It is evidence that the alternate problem described in (8) is a defensible method of choosing one of the solutions $z$ which satisfy the constraints. Csiszar has shown that the two methods of least squares and minimum I-measure are the only two which satisfy certain basic consistency requirements [Csiszar, 1989].

## A Result from Matrix Theory

$\hat{z} = A^\dagger b$ minimizes $||Az - b||$, furthermore, of all $z$ which achieve this minima $\hat{z}$ minimizes $||z||$ . Here $A^\dagger$ is the pseudo-inverse of $A$

$$A^\dagger = A^T(AA^T)^\dagger$$

Normally $(AA^T)$ is non singular, in which case $(AA^T)^\dagger = (AA^T)^{-1}$. If $(AA^T)$ is singular, then $(AA^T)^\dagger = R(R^T R)^{-1} * (F^T F)^{-1} F^T$ where the columns of $F$ are a basis for the column space of $(AA^T)$ and $R$ is defined by $(AA^T) = FR^T$ [Rao & Mitra, 1971]. This solves the inference problem described by equation (8) except for the instances where some $z_i < 0$. This has been found to be a rare occurrence. An ad-hoc solution in this case has been to fix the negative term at zero, eliminate it from all equations and solve again. If the rules are inconsistent then $\hat{z} = A^\dagger b$ will be the solution which is closest to satisfying the inconsistent constraints (in Least Mean Square).

Remember our objective is to find $p(e|x) = p(f_1, f_2, \ldots f_n|x)$. Let $s = (0\,0 \ldots 0\,1)^T$. Then

$$p(e|x) = s^T \hat{z} = s^T A^\dagger b = s^T A^T (AA^T)^{-1} b$$

Let $w^T = s^T A^T$ is the last column of $A$. Let $C = AA^T$. Then $p(e|x) = w^T C^{-1} b$. The matrix $C$ is $r$ by $r$. Both arrays have $r$ elements. Given $C$ the $p(e|x)$ can be found in $\mathcal{O}(r^3)$ operations*. This is the time complexity of matrix inversion. If a single row and column of $C$ is changed the new inverse of $C$ can be calculated in $\mathcal{O}(r^2)$ [Press et. al., 1988]. It remains to show that $C$ can be calculated in polynomial time. Remember $A$ is composed of marginals $(\alpha_1, \alpha_2, \ldots \alpha_r)^T$. We have,

$$C = AA^T = \begin{pmatrix} \alpha_1^T \alpha_1 & \alpha_1^T \alpha_2 & \ldots & \alpha_1^T \alpha_r \\ \alpha_2^T \alpha_1 & \alpha_2^T \alpha_2 & \ldots & \alpha_2^T \alpha_r \\ \vdots & \vdots & \ddots & \vdots \\ \alpha_r^T \alpha_1 & \alpha_r^T \alpha_2 & \ldots & \alpha_r^T \alpha_r \end{pmatrix}$$

These elements $(C)_{i,j}$ are known without actually multiplying the exponentially long $\alpha$ vectors. Let $m_{i,j}$ be the number of attributes common to the LHS of both rule $i$ and rule $j$. Then,

$$(C)_{i,j} = 2^{n - m_{i,i} - m_{j,j} + m_{i,j}} \tag{9}$$

Remember $n$ is the number of attributes in the entire input space. The exponent in equation (9) is the number of attributes which are not found in either rule.

---

* The $\mathcal{O}()$ notation describes an asymptotic bound:

$$f = \mathcal{O}(g) \Leftrightarrow \exists\ d, c > 0\ \forall\ n\ f(n) \leq c\ g(n) + d.$$

In words: $f(n)$ may be bounded by a linear function of $g(n)$.

## Example of Finding the C Matrix

Let $r = 3$ and $n = 3$ and let the constraints be:

$$A = \begin{pmatrix} 0 & 0 & 0 & 0 & 1 & 1 & 1 & 1 \\ 0 & 0 & 0 & 0 & 0 & 1 & 0 & 1 \\ 1 & 1 & 1 & 1 & 1 & 1 & 1 & 1 \end{pmatrix} = \begin{pmatrix} \alpha_1^T \\ \alpha_2^T \\ \alpha_3^T \end{pmatrix}$$

$$z = \begin{pmatrix} p(\overline{f}_1, \overline{f}_2, \overline{f}_3) \\ p(\overline{f}_1, \overline{f}_2, f_3) \\ p(\overline{f}_1, f_2, \overline{f}_3) \\ p(\overline{f}_1, f_2, f_3) \\ p(f_1, \overline{f}_2, \overline{f}_3) \\ p(f_1, \overline{f}_2, f_3) \\ p(f_1, f_2, \overline{f}_3) \\ p(f_1, f_2, f_3) \end{pmatrix}$$

$$b = \begin{pmatrix} p(f_1 | x) \\ p(f_1, f_3 | x) \\ 1 \end{pmatrix}$$

For this example rule 1 is $\mathbf{F_1} = f_1 \Rightarrow \mathbf{X} = x$. Rule 2 is $\mathbf{F_1} = f_1$ AND $\mathbf{F_3} = f_3 \Rightarrow \mathbf{X} = x$. Rule 3 is the constraint that every conditional on $x$ must sum to 1. Rule 1 and 3 have one term, $(F_1)$, in common term so $m_{1,2} = 1$. Rule 1 has 1 term so $m_{1,1} = 1$. By equation (9), $(C)_{1,2} = 2^{3-1-2+1} = 2$. The entire $C$ matrix is:

$$C = \begin{pmatrix} 4 & 2 & 4 \\ 2 & 2 & 2 \\ 4 & 2 & 8 \end{pmatrix}$$

Formula (9) applies only to conjunctive rules, but it can be generalized to other types of rules.

The equations described above were tested on the LED display domain. Noise was added to the 7 binary attributes describing a 7 segment LED display. These attributes were then used to predict the digit corresponding to the 7 attributes before noise was added. The rules used were simply every possible conjunction of 5 attributes. This is 672 rules, but only 21 fire for any given input. The classification procedure used was to calculate the probability of each of the 10 outputs and then choose the maximum as the predicted digit. With each attribute having a 10% chance of being changed, the classification rate was 74% which is approximately the optimum achievable with this level of noise.

## Conclusion

A method of calculating probability values from a system of marginal constraints was presented. The method uses a least squares method with the pseudo-inverse function, to collapse the exponentially long constraint equations into a matrix equation which can be solve in $\mathcal{O}(r^3)$ steps. Rules firing in an expert system may have known dependence due to common terms in the conditions which cause the rule to fire. The method presented has the effect of correcting for this dependence. The least squares method finds the solution consistent with a probability distribution which satisfies the constraints and is nearest to the equiprobable probability distribution.

The current method is appropriate for systems where all information known *apriori* may be expressed as a linear constraint on the probability distribution. Since this method finds the output probability in a predetermined number of steps, it is appropriate for calculating probabilities in a real time expert system.

**References** .

1. Peter Cheeseman, "A Method of Computing Generalize Bayesian Probability Values For Expert Systems," in IJCAI - Karlsruhe, West Germany (1983).

2. Imre Csiszar, "Why Least Squares and Maximum Entropy?" (Mathematical Institute of the Hungarian Academy of Sciences, Budapest 1989).

3. K. Konolige, Appendix D in "A Computer-Based Consultant for Mineral Exploration" SRI report Sept. (1979).

4. H.H. Ku and S. Kullback, "Approximating Discrete Probability Distributions" IEEE Trans. on Information Theory, Vol IT-15, No. 4, July (1969).

5. P.M. Lewis , "Approximation Probability Distributions to Reduce Storage Requirements." Information and Control, 2, 214-255 (1959).

6. William Press et.al. *Numerical Recipes in C*, (Cambridge Univ. Press Cambridge, 1988).

7. C.R. Rao and S.K. Mitra, *Generalize Inverse of Matrices and its Applications* (Wiley and Sons New York, 1971).